\documentclass[review]{elsarticle}

\usepackage{color}
\usepackage{longtable}
\usepackage{caption}
\usepackage{lineno,hyperref}
\usepackage{multirow}
\usepackage{verbatim}
\usepackage{amsmath}
\usepackage{makecell}
\usepackage[noend]{algpseudocode}
\usepackage{algorithmicx,algorithm}
\modulolinenumbers[5]

\journal{Pattern Recognition}

\newcommand{\tabincell}[2]{\begin{tabular}{@{}#1@{}}#2\end{tabular}}
\newcommand{\eg}{\emph{e.g. }}
\newcommand{ \etal}{\emph{et.al.\ }}
\newcommand{\ie}{\emph{i.e. }}
\newcommand{\etc}{\emph{etc}}

\usepackage{amssymb}
\usepackage[flushleft]{threeparttable}

\begin{document}

\begin{frontmatter}

\title{Binary Neural Networks: A Survey}

\author{Haotong~Qin\fnref{1}}
\author{~Ruihao~Gong\fnref{1}}
\author{~Xianglong~Liu\fnref{1,2}\footnote{Corresponding author}}
\author{~Xiao~Bai\fnref{5}}
\author{~Jingkuan~Song\fnref{3}}
\author{~Nicu~Sebe\fnref{4}}
\address[1]{State Key Lab of Software Development Environment, Beihang University, Beijing, China.}
\address[2]{Beijing Advanced Innovation Center for Big Data-Based Precision Medicine, Beihang University, Beijing, China.}
\address[3]{Center for Future Media and School of Computer Science and Engineering, University of Electronic Science and Technology of China, Chengdu, China.}
\address[4]{Department of Information Engineering and Computer Science, University of Trento, Trento, Italy.}
\address[5]{School of Computer Science and Engineering, Beijing Advanced Innovation Center for Big Data and Brain Computing, Jiangxi Research Institute, Beihang University, Beijing, China}

\begin{abstract}
The binary neural network, largely saving the storage and computation, serves as a promising technique for deploying deep models on resource-limited devices. However, the binarization inevitably causes severe information loss, and even worse, its discontinuity brings difficulty to the optimization of the deep network. To address these issues, a variety of algorithms have been proposed, and achieved satisfying progress in recent years. In this paper, we present a comprehensive survey of these algorithms, mainly categorized into the native solutions directly conducting binarization, and the optimized ones using techniques like minimizing the quantization error, improving the network loss function, and reducing the gradient error. We also investigate other practical aspects of binary neural networks such as the hardware-friendly design and the training tricks. Then, we give the evaluation and discussions on different tasks, including image classification, object detection and semantic segmentation. Finally, the challenges that may be faced in future research are prospected.
\end{abstract}

\begin{keyword}
binary neural network\sep deep learning\sep model compression\sep network quantization\sep model acceleration
\end{keyword}

\end{frontmatter}


\section{Introduction}
\label{section1}

With the continuous development of deep learning~\cite{Lecun2015Deep}, deep neural networks have made significant progress in various fields, such as computer vision, natural language processing and speech recognition. {Convolutional neural networks (CNNs) have been proved to be reliable in the fields of image classification~\cite{7298594,NIPS2015_5638,DBLP:journals/pr/NogueiraPS17,DBLP:journals/pr/LiuFGWP17,DBLP:journals/pr/DengLLT18}, object detection~\cite{DBLP:journals/corr/Girshick15,DBLP:journals/corr/abs-1904-02701,ge2017DetectingMasked,DBLP:journals/pr/LopesASO17} and object recognition~\cite{krizhevsky2012imagenet,simonyan2015very,7298594,DBLP:journals/pr/WuWGL18,DBLP:journals/pr/LiLZW20}, and thus have been widely used in practice.} 

Owing to the deep structure with a number of layers and millions of parameters, the deep CNNs enjoy strong learning capacity, and thus usually achieve satisfactory performance. For example, the VGG-16~\cite{simonyan2015very} network contains about 140 million 32-bit floating-point parameters, and can achieve 92.7\% top-5 test accuracy for image classification task on ImageNet dataset. The entire network needs to occupy more than 500 megabytes of storage space and perform $1.6\times{10}^{10}$ floating-point arithmetic operations. This fact makes the deep CNNs heavily rely on the high-performance hardware such as GPU, while in the real-world applications, usually only the devices (\eg, the mobile phones and embedded devices) with limited computational resources are available~\cite{cheng2017survey}. For example, embedded devices based on FPGAs usually have only a few thousands of computing units, far from dealing with millions of floating-point operations in the common deep models. There exists a severe contradiction between the complex model and the limited computational resources. Although at present, a large amount of dedicated hardware emerges for deep learning~\cite{chen2018eyeriss,Sze2017Hardware,Chen2016Eyeriss,Chen201614,EyerissAnEnergy-Efficient}, providing efficient vector operations to enable fast convolution in forward inference, the heavy computation and storage still inevitably limit the applications of the deep CNNs in practice. {Besides, due to the huge model parameter space, the prediction of the neural networks is usually viewed as a black-box, which brings great challenges to the interpretability of CNNs. Some works like~\cite{Zhang2019InterpretingAI,Yu2019TowardsNN,Liu2019TrainingRD} empirically explore the function of each layer in the network. They visualize the feature maps extracted by different filters and view each filter as a visual unit focusing on different visual components.}

{From the aspect of explainable machine learning, we can summarize that some filters are playing a similar role in the model, especially when the model size is large. So it is reasonable to prune some useless filters or reduce their precision to lower bits. On the one hand, we can enjoy more efficient inference with such compression technique. On the other hand, we can utilize it to further study the interpretability of CNNs, \ie, finding out which layer is important, which layer is useless and can be removed from the black-box, what structure is beneficial for accurate prediction.}
Many prior studies have proven that there usually exists large redundancy in the deep structure~\cite{Izui1990Analysis,cheng2015an,han2015learning,Srinivas2015Data}. For example, by simply discarding the redundant weights, one can keep the performance of the ResNet-50~\cite{he2016deep}, and meanwhile save more than 75\% of parameters and 50\% computational time. {In the literature, approaches for compressing the deep networks can be classified into five categories: parameter pruning~\cite{han2015learning, han2016deep, he2017channel,ge2017Compressing}, parameter quantizing~\cite{gong2014compressing,wu2016quantized,Vanhoucke2011ImprovingTS,gupta2015deep,ge2018EfficientDeepLearning,DBLP:conf/mm/ZhaoH0H18,DBLP:conf/eccv/HuLWZC18,DBLP:conf/nips/ChenWP19,Wu2020Rotation,zhu2019unified}, low-rank parameter factorization~\cite{denton2014exploiting,lebedev2015speeding,jaderberg2014speeding,lebedev2016fast,DBLP:journals/pr/WenZXYH18}, transferred/compact convolutional filters~\cite{mobilenet,mobilenet_v2,shufflenet,shufflenet_v2}, and knowledge distillation~\cite{hinton2015distilling,xu2018training,chen2018darkrank,Yim2017A,zagoruyko2017paying,DBLP:journals/pr/DingCH19}.} The parameter pruning and quantizing mainly focus on eliminating the redundancy in the model parameters respectively by removing the redundant/uncritical ones or compressing the parameter space (\eg, from the floating-point weights to the integer ones). Low-rank factorization applies the matrix/tensor decomposition techniques to estimate the informative parameters using the proxy ones of small size. The compact convolutional filter based approaches rely on the carefully-designed structural convolutional filters to reduce the storage and computation complexity. The knowledge distillation methods try to distill a more compact model to reproduce the output of a larger network.

Among the existing network compression techniques, quantization based one serves as a promising and fast solution that yields highly compact models compared to their floating-point counterparts, by representing the network weights with very low precision. Along this direction, the most extreme quantization is binarization, the interest in this survey. Binarization is a 1-bit quantization where data can only have two possible values, namely -1(0) or +1. For network compression, both the weight and activation can be represented by 1-bit without taking too much memory. Besides, with the binarization, the heavy matrix multiplication operations can be replaced with light-weighted bitwise XNOR operations and Bitcount operations. Therefore, compared with other compression methods, binary neural networks enjoy a number of hardware-friendly properties including memory saving, power efficiency and significant acceleration. The pioneering work like BNN~\cite{BNN} and XNOR-Net~\cite{XNOR-Net} has proven the effectiveness of the binarization, namely, up to $32\times$ memory saving and $58\times$ speedup on CPUs, which has been achieved by XNOR-Net for a 1-bit convolution layer. Following the paradigm of binary neural network, in the past years a large amount of research has been attracted on this topic from the fields of computer vision and machine learning~\cite{Lecun2015Deep,7298594,simonyan2015very,he2016deep}, and has been applied to various popular tasks such as image classification\cite{BinaryConnect,DoReFa-Net,LQ-Net,Bi-Real,Gong:iccv19}, detection~\cite{BWBDN,Li_2019_CVPR}, and so on. {With the binarization technique, the importance of a layer can be easily validated by switching it to full-precision or 1-bit. If the performance greatly decreases after binarizing certain layer, we can conclude that this layer is on the critical path of the network. Furthermore, it is also significant to find out whether the full-precision model and the binarized model work in the same way from the explainable machine learning view.}

{Besides focusing on the strategies of model binarization, many studies have attempted to reveal the behaviors of model binarization, and further explain the connections between the model robustness and the structure of deep neural networks. This possibly helps to approach the answers to the essential questions: how does the deep network work indeed and what network structure is better? It is very interesting and important to well investigate the studies of binary neural network, which will be very beneficial for understanding the behaviors and structures of the efficient and robust deep learning models. Some of studies in the literature have shown that binary neural networks can filter the input noise, and pointed out that specially designed BNNs are more robust compared with the full-precision neural networks. \cite{lin2018defensive} shows that noise is continuously amplified during the forward propagation of neural networks, and binarization improves robustness by keeping the magnitude of the noise small.}

{The studies based on BNNs can also help us to analyze how structures in deep neural networks work. Liu \etal creatively proposed Bi-Real Net, which added additional shortcuts (Bi-Real) to reduce the information loss caused by binarization~\cite{Bi-Real}. This structure works like the shortcut in ResNet and it helps to explain why the widely used shortcuts can improve performance of deep neural networks to some extent. On the one hand, by visualizing the activations, it can be seen that more detailed information in the shallow layer can be passed to the deeper layer during forward propagation. On the other hand, gradients can be directly backward propagated through the shortcut to avoid gradient vanish problem.
Zhu \etal leveraged ensemble methods to improve the performance of BNNs by building several groups of weak classifiers, and the ensemble methods improve the performance of BNNs although sometimes face over-fitting problem~\cite{BENN}. Based on analysis and experimentation of BNNs, they showed that the number of neurons is more important than the bit-width and it may not be necessary to use real-valued neurons in deep neural networks, which is similar to the principle of biological neural networks.
Besides, reducing the bit-width of certain layer to explore its effect on accuracy is one effective approach to study the interpretability of deep neural networks. There are many works to explore the sensitivity of different layers to binarization. It is a common sense that the first layer and the last layer should be kept in higher precision, which means that these layers play a more important role in the prediction of neural networks.}

This survey tries to exploit the nature of binary neural networks and categorizes the them into the naive binarization without optimizing the quantization function and the optimized binarization including minimizing quantization error, improving the loss function, and reducing the gradient error. It also discusses the hardware-friendly methods and the useful tricks of training binary neural networks. In addition, we present the common datasets and network structures of evaluation, and compare the performance of current methods on different tasks. 
The organization of the remaining part is given as the following. Section \ref{section2} introduces the preliminaries for binary neural network. Section \ref{section3} presents the existing methods falling in different categories and lists the training tricks in practice. Section \ref{section4} gives the evaluation protocols and performance analysis. Finally, we conclude and point out the future research trends in Section \ref{section5}.

\section{Preliminary}
\label{section2}

In full-precision convolutional neural networks, the basic operation can be expressed as 
\begin{equation}
{\mathbf z=\sigma(\mathbf w\otimes\mathbf a)}
\end{equation}
{where $\mathbf w$ and $\mathbf a$ represent the weight tensor and activation tensor generated by the previous network layer, respectively. $\sigma(\cdot)$ is the non-linear function and $\mathbf z$ is the output tensor and $\otimes$ represents the convolution operation. In the forward inference process of neural networks, the convolution operation contains a large number of floating-point operations, including floating-point multiplication and floating-point addition, which correspond to the vast majority of calculations in neural network inference.}

\subsection{Forward Propagation}
The goal of network binarization is to represent the floating-point weights $\mathbf w$ and/or activations $\mathbf a$ using 1-bit. The popular definition of the binarization function is given as follows:
\begin{equation}
Q_w(\mathbf w)=\alpha\mathbf{b_w},\quad Q_a(\mathbf a)=\beta\mathbf{b_a}
\end{equation}
{where $\mathbf b_{\mathbf w}$ and $\mathbf b_{\mathbf a}$ are the tensor of binary weights (kernel) and binary activations, with the corresponding scalars $\alpha$ and $\beta$. In the literature, the $\mathtt{sign}$ function is widely used for $Q_w$ and $Q_a$:}
\begin{equation}
\mathtt{sign}(x)=
\left\{\begin{array}{ll}{+1,} & {\text {\rm if }\ x \ge 0} \\
{-1,} & {\text {\rm otherwise }}\end{array}\right.
\end{equation}

\begin{figure}[htbp]
	\begin{center}
		\includegraphics[width=1\linewidth]{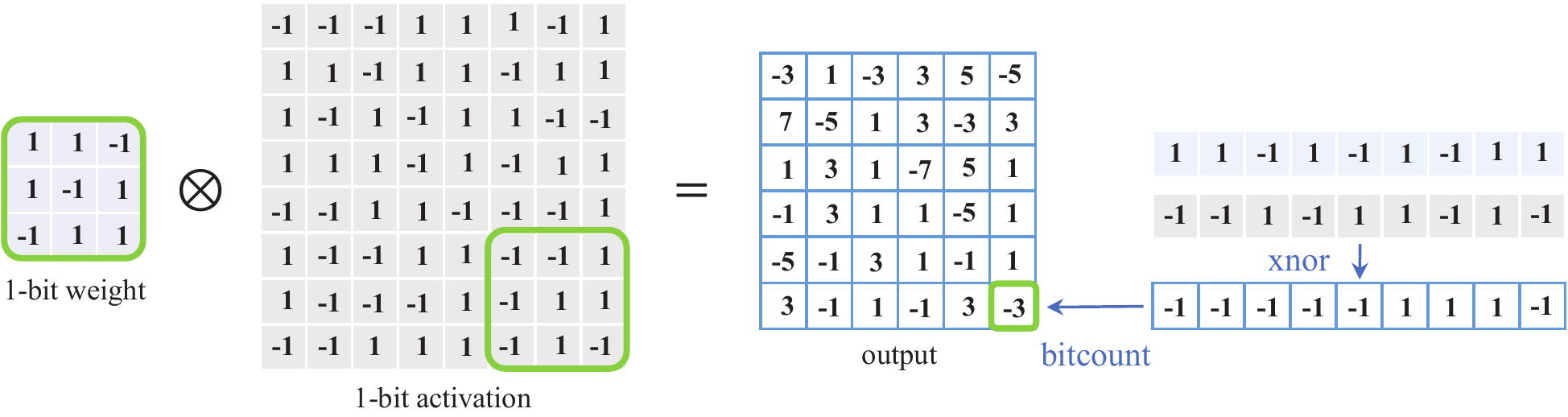}
	\end{center}
	\vspace{-0.2in}
	\caption{Convolution Process of Binary Neural Networks}
	\label{fig:bi-conv}
	\vspace{-0.1in}
\end{figure}

With the binarized weights and activations, the vector multiplication in forward propagation can be reformulated as
\begin{equation}
{\mathbf{z} = \sigma(Q_w(\mathbf w) \otimes Q_a(\mathbf a))= \sigma({{\alpha}}{{\beta}}({\mathbf{b_w}}\odot {\mathbf{b_a}})),}
\end{equation} 
where $\odot$ denotes the inner product for vectors with bitwise operation XNOR-Bitcount. Figure \ref{fig:bi-conv} shows the convolution process in the binary neural networks.

\subsection{Backward Propagation}
Similar to training a full-precision neural network model, when training a binary neural network, it is still straightforward to adopt the powerful backward propagation (BP) algorithm based on the gradient descent to update the parameters.
However, usually the binarization function (\eg, $\mathtt{sign}$) is not differentiable, and even worse, the derivative value in part of the function vanishes (\eg, 0 almost everywhere for $\mathtt{sign}$). Therefore, the common gradient descent based BP algorithm cannot be directly applied to update the binary weights.

{Fortunately, the technique called straight-through estimator (STE) has been proposed by Hinton \etal to address the gradient problem occurring when training deep networks binarized by $\mathtt{sign}$ function~\cite{STE}.
The function of STE is defined as follows}
\begin{equation}
\mathtt{clip}(x,-1,1)=\max (-1, \min (1, x)).
\end{equation}
Through STE, the binary neural network can be directly trained using the same gradient descent method as the ordinary full-precision neural network.
{However, when the $\mathtt{clip}$ function is used in backward propagation, if the absolute value of full-precision activations are greater than 1, it cannot be updated in backward propagation. Therefore, in the practical scenarios, the $\mathtt{Identity}$ function is also chosen to approximate the derivative of the $\mathtt{sign}$ function.}

\section{Binary Neural Networks}
\label{section3}

Compared with the full-precision neural network, the binary neural networks based on 1-bit representation replace the floating-point multiplication and addition operations by the efficient XNOR-Bitcount operations, and thus largely reduce the storage space and the inference time. However, the binarization of weights and activations will cause a severe deviation from the full-precision ones. {Also, as aforementioned, the discrete binarization makes the popular gradient descent based BP algorithm usually fail to pursue the satisfactory solution, even with the STE technique.} Therefore, the binary neural networks inevitably suffer from the performance degradation. It is still an open research problem that how to optimize the binary neural network.

{In recent years, a variety of binary neural networks have been proposed, from the native solutions that directly binarize the weights and inputs using the pre-defined binarization function, to the optimization based ones using different techniques that treat the problem from different perspectives: approximate the full-precision values by minimizing the quantization error, constrain the weights by modifying the network loss function, and learn the discrete parameters by reducing the gradient error. Table \ref{tab:over} summaries the surveyed binarization methods in different categories.}

\begin{table}[]
\setlength{\abovecaptionskip}{0.cm}
\caption{{Overview of Binary Neural Networks}}\label{tab:over}
\scriptsize
\centering
\begin{threeparttable}
\setlength{\tabcolsep}{0.5mm}{
\begin{tabular}{|c|c|c|c|c|c|c|c|}
\hline
\multicolumn{2}{|c|}{\multirow{2}{*}{\textbf{Type}}} & \multirow{2}{*}{\textbf{Method}} & \multirow{2}{*}{\textbf{Key Tech.}}
     & \multicolumn{4}{|c|}{\textbf{Tricks}} \\ \cline{5-8} 

\multicolumn{2}{|c|}{}         & & & ST & OP & AQ & GA\\ \cline{1-8}
     
\multicolumn{2}{|c|}{\multirow{3}{*}{\tabincell{c}{Naive Binary\\Neural Networks}}}  & BinaryConnect~\cite{BinaryConnect}                               &\multirow{3}{*}{\tabincell{c}{FP:\quad $\mathtt{sign}(x)$\\BP:\quad STE}}               &  - & \tiny{A} & - & - \\ \cline{3-3} \cline{5-8}

\multicolumn{2}{|c|}{}         & Bitwise Neural Networks~\cite{BitwiseNN}             &                & - & - & - & -      \\ \cline{3-3} \cline{5-8}

\multicolumn{2}{|c|}{}         & Binarized Neural Networks~\cite{BNN}                 &                & - & \tiny{AM} & - & -      \\ \cline{1-8}

\multirow{32}{*}{\tabincell{c}{{Optimization}\\\\{Based}\\\\{Binary}\\\\{Neural}\\\\{Networks}}} & \multirow{14}{*}{\tabincell{c}{Minimize\\the\\Quantization\\Error}}  & Binary Weight Networks~\cite{BNN}  &

\multirow{14}{*}{\tabincell{c}{$J(\mathbf{b}, \alpha)=$\\ $\Vert\mathbf{x}-\alpha \mathbf{b}\Vert^{2}$ \\\\ $\alpha^{*},
\mathbf{b}^{*}=$\\ $\mathop{\arg\min}\limits_{\alpha, \mathbf{b}}{J(\mathbf{b},
\alpha)}$}}
& - & \tiny{S} & - & -\\ \cline{3-3} \cline{5-8}

&         & XNOR-Net~\cite{XNOR-Net}                            &                & \tiny{RB+RP} & \tiny{A} & - & - \\ \cline{3-3} \cline{5-8}

&         & DoReFa-Net~\cite{DoReFa-Net}                        &                & - & \tiny{A} & - & - \\ \cline{3-3} \cline{5-8}

&         & High-Order Residual Quantization~\cite{HORQ}        &                & - & \tiny{A} & - & - \\ \cline{3-3} \cline{5-8}

&         & ABC-Net~\cite{ABC-Net}                              &                & - & \tiny{S} & - & - \\ \cline{3-3} \cline{5-8}

&         & Two-Step Quantization~\cite{TSQ}                    &                & \tiny{RB} & - & - & -      \\ \cline{3-3} \cline{5-8}

&         & Binary Weight Networks via Hashing\cite{BWN-Hashing}&                & - & \tiny{S} & - & - \\ \cline{3-3} \cline{5-8}

&         & PArameterized Clipping acTivation~\cite{PACT}       &                & - & \tiny{A} & - & - \\ \cline{3-3} \cline{5-8}

&         & LQ-Nets~\cite{LQ-Net}                               &                & \tiny{RB} & - & - & - \\ \cline{3-3} \cline{5-8}

&         & Wide Reduced-Precision Networks~\cite{WRPN}         &                & \tiny{WD} & \tiny{A} & - & - \\ \cline{3-3} \cline{5-8}

&         & {XNOR-Net++}~\cite{Bulat2019XNORNetIB}         &                & {-} & {\tiny{A}} & {-} & {-} \\ \cline{3-3} \cline{5-8}

&         & Learning Symmetric Quantization~\cite{SYQ}           &               & - & - &\checkmark& -  \\ \cline{3-3} \cline{5-8} 

&         & {BBG~\cite{DBLP:journals/corr/abs-1909-12117} }          &               & {SC} & {-} &{-}& {-} \\ \cline{3-3} \cline{5-8} 

&         & {Real-to-Bin~\cite{martinez2020training} }          &               & {SC} & {A} &{-}& {\checkmark} \\ \cline{2-8} 

&  \multirow{7}{*}{\tabincell{c}{Improve\\Network\\Loss\\Function}} & Distilled Binary Neural Network~\cite{DistilledBNN}                    &\multirow{7}{*}{\tabincell{c}{$\mathcal{L}_{\text {\rm total}}^{b}=$\\$\mathcal{L}_{\rm original}^{b}+$\\ $\lambda \mathcal{L}_{\rm Customized}^{b}$}}                & - & \tiny{S} & - & -     \\ \cline{3-3} \cline{5-8}

&         & Distillation and Quantization~\cite{Distillation-Quant} &            & - & \tiny{S} & - & - \\ \cline{3-3} \cline{5-8}

&         & Apprentice~\cite{Apprentice}                                 &       & - & - & - & -  \\ \cline{3-3} \cline{5-8}

&         & Loss-Aware Binarization~\cite{Loss-Aware-BNN}           &            & - & \tiny{A} & - & - \\ \cline{3-3} \cline{5-8}

&         & Incremental Network Quantization~\cite{INQ}           &              & - & \tiny{S} &\checkmark& - \\ \cline{3-3} \cline{5-8}

&         & BNN-DL~\cite{Regularize-act-distribution} & & - & \tiny{R} & - &\checkmark \\ \cline{3-3} \cline{5-8}

&         & {CI-BCNN~\cite{LearningChannel-Wise}} & & {-} & {\tiny{R}} & {-} &{\checkmark} \\ \cline{3-3} \cline{5-8}

&         & Main/Subsidiary Network~\cite{Subsidiary} & & \tiny{RB} & - & - & - \\ \cline{3-3} \cline{2-8}

&\multirow{11}{*}{\tabincell{c}{Reduce\\the\\Gradient\\Error}} & Bi-Real Net~\cite{Bi-Real}                                &\multirow{11}{*}{\tabincell{c}{Customized\\$\mathtt{ApproxFunc}$ (FP)\\or\\$\mathtt{QuantFunc}$ (BP)\\or\\$\mathtt{UpdateFunc}$ (BP)}}                 &\tiny{SC} & \tiny{S} & - &\checkmark      \\ \cline{3-3} \cline{5-8}

&         & Circulant Binary Convolutional Networks\cite{CirculantBNN}    &      &\tiny{SC} & \tiny{S} & - &\checkmark       \\ \cline{3-3} \cline{5-8}

&         & Half-wave Gaussian Quantization~\cite{HWGQ}            &             & \tiny{RB} & \tiny{S} & - &\checkmark \\ \cline{3-3} \cline{5-8}

&         & BNN+~\cite{BNN+}                                       &             & \tiny{RB} & \tiny{A} & - &\checkmark \\ \cline{3-3} \cline{5-8}

&         & Differentiable Soft Quantization~\cite{Gong:iccv19}           &          & - & \tiny{A} & - &\checkmark \\ \cline{3-3} \cline{5-8}

&         & BCGD~\cite{BCGD}           &          & - & - & - &\checkmark \\ \cline{3-3} \cline{5-8}

&         & ProxQuant~\cite{proxquant}           &          & - & \tiny{A} & - &\checkmark \\ \cline{3-3} \cline{5-8}

&         & Quantization Networks~\cite{quantization_networks}           &          & - & \tiny{S} & - &\checkmark \\ \cline{3-3} \cline{5-8}

&         & {Self-Binarizing Networks~\cite{selfBN}}           &          & {-} & {\tiny{A}} & {-} &{\checkmark} \\ \cline{3-3} \cline{5-8}

&         & {Improved Training BNN~\cite{ImprovedTraining}}           &          & {-} & {\tiny{A}} & {-} & {\checkmark} \\ \cline{3-3} \cline{5-8}

&         & {IR-Net~\cite{IRNet}}           &          & {-} & {\tiny{S}} & {\checkmark} & {\checkmark} \\

\hline
\end{tabular}}
\begin{tablenotes}
    \footnotesize
    \item[*] Tech. = Technology. Tricks: ST = Structure Transformation, OP = Optimizer, AQ = Asymptotic Quantization, GA = Gradient Approximation. Optimizer: S = SGD, A = Adam, AM = AdaMax, R = RMSprop. Structure Transformation: RB = Reorder BN layer, RP = Reorder Pooling layer, WD = Widen, SC = Shortcut. FP = Forward Propagation, BP = Backward Propagation.
\end{tablenotes}
\end{threeparttable}
\end{table} 

\subsection{Naive Binary Neural Networks}
The naive binary neural networks directly quantize the weights and activations in the neural network to 1-bit by the fixed binarization function. {Then the basic backward propagation strategy equipped with STE is applied to optimize the deep models in the standard training way.}

In 2016 Courbariaux \etal proposed BinaryConnect~\cite{BinaryConnect} that pioneered the study of binary neural networks. BinaryConnect converts the full-precision weights inside the neural network into 1-bit binary weights. In the forward propagation of training, a stochastic binarization method is adopted to quantize the weights, and the effect of the binary weights during inference is simulated. During the backward propagation, a $\mathtt{clip}$ function is introduced to cut off the update range of the full-precision weights to prevent the real-valued weights from growing too large without any impact on the binary weights. Though after model binarization the parameters of the neural network model is greatly compressed (even with large quantization error), the binary model can closely reach the state-of-the-art performance on some datasets in the image classification tasks. The stochastic binarization method in BinaryConnect is defined as:
\begin{equation}
w_b =
\left\{\begin{array}{ll}{+1,} & {{\rm with\ probability }\ p=\hat{\sigma}(w)} \\
{-1,} & {{\rm with\ probability }\ 1-p}\end{array}\right.
\end{equation}
where $\hat{\sigma}$ is the ``hard sigmoid" function:
\begin{equation}
    \hat{\sigma}(x) =\mathtt{clip}(\frac{x+1}{2}, 0, 1) = \max(0, \min(1, \frac{x+1}{2}))
\end{equation}

Following the paradigm of binarizing the network, Courbariaux \etal further introduced Binarized Neural Network (BNN)~\cite{BNN}, presenting the training and acceleration skills in detail. It proved the practicability and acceleration capability of binary neural networks from both theoretical and practical aspects. For the inference acceleration of networks with batch normalization, this method also devised techniques like Shift-based Batch Normalization and XNOR-Bitcount. The experiments on image classification show that BNN takes $32\times$ less storage space and 60\% less time. Smaragdis \etal also studied the network binarization and developed Bitwise Neural Network especially suitable for resource-constrained environments~\cite{BitwiseNN}.

\subsection{{Optimization Based Binary Neural Networks}}
The naive binarization methods own the advantages of saving computational resources by quantizing the network in a very simply way. However, without considering the effect of the binarization in the forward and backward process, these methods inevitably suffer the accuracy loss for the wide tasks. Therefore, in order to mitigate the accuracy loss in the binary neural network, in the past years, a great number of optimization-based solutions have been proposed and shown the successful improvement over the native ones.

\subsubsection{Minimize the Quantization Error}
For the optimization of binary neural networks, a common practice is to reduce the quantization error of weight and activation. This is a straightforward solution similar to the standard quantization mechanism that the quantized parameter should approximate the full-precision parameter as closely as possible, expecting that the performance of the binary neural network model will be close to the full-precision one.

\begin{figure}[htbp]
	\begin{center}
		\includegraphics[width=1\linewidth]{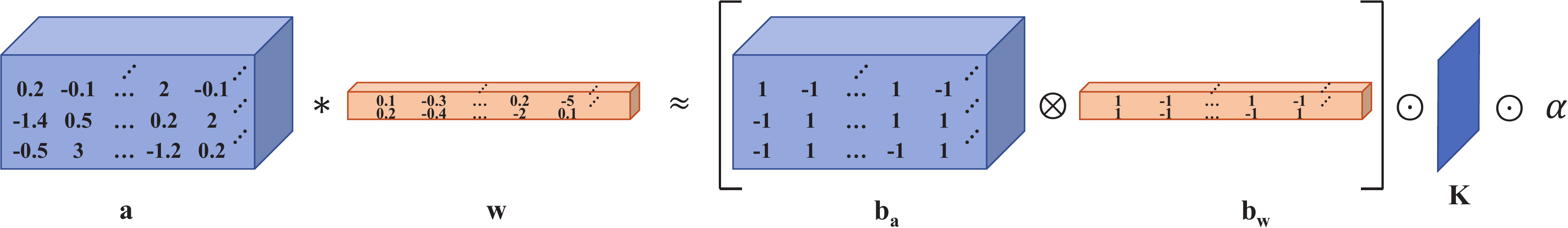}
	\end{center}
	\vspace{-0.2in}
	\caption{Binarization and Convolution Process of XNOR-Net}
	\label{fig:xnor}
	\vspace{-0.1in}
\end{figure}

As the early research considering the quantization error, Rastegari \etal proposed Binary Weight Networks (BWN) and XNOR-Net~\cite{XNOR-Net}. BWN adopts the setting of binary weights and full-precision activations, while XNOR-Net binarizes both weights and activations. Different from the prior studies,~\cite{XNOR-Net} well approximates the floating-point parameters by introducing a scaling factor for the binary parameter. Specifically, the weight quantization process in BWN and XNOR-Net can be formulated as $\mathbf w\approx\alpha\mathbf{b_w}$, where $\alpha$ is the floating-point scaling factor for the binarized weight $\mathbf{b_w}$. This means that the weights in BWN are binarized to $\{-\alpha, +\alpha\}$, but still can bring the benefits of fast computation. {Then minimizing the quantization error can help to find the optimal scaling factor and binary parameters:}
\begin{equation}
{\mathop{\min}\limits_{\alpha, \mathbf{b_{\mathbf w}}}\|\mathbf{w}-\alpha \mathbf{b_{\mathbf w}}\|^{2}}
\end{equation}
The solution enjoys much less quantization error than directly using 1-bit (-1/+1), thereby improving the inference accuracy of the network. Figure \ref{fig:xnor} shows the binarization and the corresponding convolution process in XNOR-Net. Similar idea was also proposed in Binary Weight Networks via Hashing (BWNH)~\cite{BWN-Hashing}, which considers the quantizing process as a hash map with scaling factors. The DoReFa-Net~\cite{DoReFa-Net} further extends XNOR-Net, so that the network training can be accelerated using quantized gradients. Mishra \etal devised Wide Reduced-Precision Networks (WRPN)~\cite{WRPN} that also minimize the quantization error in a similar way to XNOR-Net, but increase the number of filters in each layer. Compared with directly binarizing the network, widening and binarizing together can achieve a good balance between the precision and the acceleration. The work of Faraone \etal groups parameters in training process and gradually quantizes each group with optimized scaling factor to minimize the quantization error~\cite{SYQ}.

To further reduce the quantization error, High-Order Residual Quantization (HORQ)~\cite{HORQ} adopts a recursive approximation to the full-precision activation based on the quantized residual, instead of one-step approximation used in XNOR-Net. It generates the final quantized activation by a linear combination of the approximation in each recursive step. In a very similar way, Lin \etal designed ABC-Net~\cite{ABC-Net} that linearly combines multiple binary weight matrices and scaling factors to fit the full-precision weights and activations, which can largely reduce the information loss caused by binarization. Wang \etal pointed out the shortcoming of the previous methods that separately minimizing the quantization error of weights and activations can hardly promise the outputs to be similar to the full-precision ones~\cite{TSQ}. To address this problem, a two-step quantization (TSQ) method is designed. During the first step, all weights are full-precision values and all activations are quantized into low-bit format with a learnable quantization function $Q_a$. During the second step, $Q_a$ is fixed and the low-bit weight vector $\mathbf{b_w}$ and scaling factor $\alpha$ are learned as follow:
\begin{equation}
{\min\limits_{\alpha, \mathbf{\mathbf{b_w}}} \quad\left\|\mathbf z-Q_{a}\left(\alpha(\mathbf{a}\odot \mathbf{b_w})\right)\right\|_{2}^{2},}
\end{equation}
which can be solved efficiently in an iterative manner.

The aforementioned methods usually choose the fixed binarization function (\eg, $\mathtt{sign}$ function). One can also adopt more flexible binarization function and learn its parameters during minimizing the quantization error. To achieve this goal, Choi \etal proposed PArameterized Clipping Activation (PACT)~\cite{PACT} with a learnable upper bound for the activation function. The optimized upper bound of each layer is able to ensure that the quantization range of each layer is aligned with the original distribution. In practice, PACT performs better on binary networks, and can achieve accuracy close to full-precision network on larger networks. In~\cite{LQ-Net}, the Learned Quantization (LQ-Nets) attempts to minimize quantization error by jointly training neural networks and quantizers in the network. Different from the previous work, LQ-Nets learn the quantization thresholds and cutoff values by minimizing the quantization error during the network training, and can support arbitrary bit quantization. In~\cite{Xu2019Accurate}, trainable scaling factors for both weights and activations are introduced to increase the value range. {And based on XNOR-Net, Bulat \etal fused the activation and weight scaling factors into a single one that is learned discriminatively via backward propagation and proposed XNOR-Net++~\cite{Bulat2019XNORNetIB}.}

\subsubsection{Improve the Network Loss Function}

Minimizing the quantization error tries to retain the values of full-precision weights and activations, and thus reduces the information loss in each layer. However, only focusing on the local layers can hardly promise the exact final output passed through a series of layers. Therefore, it is highly required that the network training can globally take the binarization as well as the task-specific objective into account. Recently, an amount of research works at finding the desired network loss function that can guide the learning of the network parameters with restrictions brought by binarization.

Usually the general binarization scheme only focuses on accurate local approximation of the floating-point values and ignores the effect of binary parameters on the global loss. In~\cite{Loss-Aware-BNN}, Hou \etal proposed Loss-Aware Binarization (LAB) that directly minimizes the overall loss associated with binary weights using the quasi-Newton algorithm. The method utilizes information from the second-order moving average that has been calculated by the Adam optimizer to find optimal weights with consideration of the characteristics of binarization. Apart from considering the task-relevant loss from a quantization view, devising additional quantization-aware loss item is proved to be practical. In~\cite{Regularize-act-distribution}, Ding \etal summarized the problems caused by forward binarization and backward propagation in binary neural networks, including ``degeneration", ``saturation" and ``gradient mismatch". To address these issues, a distribution loss was introduced to explicitly regularize the activation distribution as follows: 
\begin{equation}
\mathcal{L}_{total}=\mathcal{L}_{CE}+\lambda \mathcal{L}_{DL}
\end{equation}
where $\mathcal{L}_{CE}$ is the common cross-entropy loss for training deep neural networks, $\mathcal{L}_{DL}$ is the distribution loss for learning the proper binarization, and $\lambda$ balances the effect of the two types of losses. With the guide of additional loss, the learned neural network can effectively avoid the aforementioned obstacles and is friendly to binarization. The Incremental Network Quantization (INQ) method~\cite{INQ} proposed by Zhou \etal also proved this point, which adds a regularization term in loss function.
\begin{figure}[htbp]
	\begin{center}
		\includegraphics[width=1\linewidth]{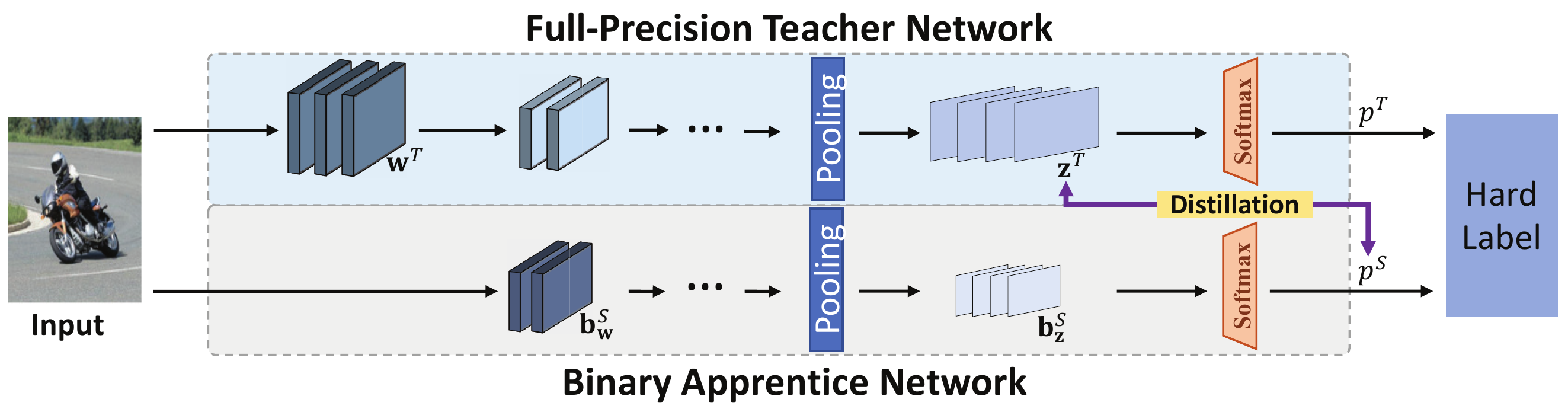}
	\end{center}
	\vspace{-0.2in}
	\caption{Schematic of Apprentice Network}
	\label{fig:apprentice}
	\vspace{-0.1in}
\end{figure}

The guiding information for training accurate binary neural networks can also derive from the knowledge of a large full-precision model. The Apprentice method~\cite{Apprentice} trains a low-precision student network using a well-trained, full-precision, large-scale teacher network, using the following loss function:
\begin{equation}
\mathcal{L}\left(x ; \mathbf {w}^{T}, \mathbf{b}_\mathbf{w}^{S}\right)=\alpha \mathcal{H}\left(y, p^{T}\right)+\beta \mathcal{H}\left(y, p^{S}\right)+\gamma \mathcal{H}\left(z^{T}, p^{S}\right)
\end{equation}
where $\mathbf {w}^T$ and $\mathbf {b}_\mathbf{w}^S$ are the full-precision weights of the teacher model and binary weights of the student (apprentice) model respectively, $y$ is the label for sample $x$, $\mathcal{H}(\cdot)$ is the soft and hard label loss function between the teacher and apprentice model, and $\alpha,\ \beta,\ \gamma$ are the weighting factors, $p^T$ and $p^S$ are the predictions of the teacher and student model, respectively. Under the supervision of the teacher network, the binary network can preserve the learning capability and thus obtain the close performance to the teacher network. The process of knowledge distillation is shown in Figure \ref{fig:apprentice}. Similar mimic solutions like Distillation and Quantization (DQ)~\cite{Distillation-Quant}, Distilled Binary Neural Network (DBNN)~\cite{DistilledBNN} and Main/Subsidiary Network~\cite{Subsidiary} have been studied, and their experiments demonstrate that the loss functions related to the full-precision teacher model help to stabilize the training of binary student model with high accuracy. CI-BCNN proposed in~\cite{LearningChannel-Wise} mines the channel-wise interactions, through which prior knowledge is provided to alleviate inconsistency of signs in binary feature maps and preserves the information of input samples during inference. {\cite{martinez2020training} built strong BNNs with a loss function during training, which matches the spatial attention maps computed at the output of the binary and real-valued convolutions.}

\subsubsection{Reduce the Gradient Error}
Training of binary neural networks still relies on the popular BP algorithm. To deal with the gradients for the non-differential binarization function, straight-through estimator (STE) technique is often adopted to estimate the gradients in backward propagation~\cite{STE}. However, there exists obvious gradient mismatch between the gradient of the binarization function (\eg, $\mathtt{sign}$) and STE (\eg, $\mathtt{clip}$). Besides, it also suffers the problem that the parameters outside the range of $[-1, +1]$ will not be updated. These problems easily lead to the under-optimized binary networks with severe performance degradation.  

Intuitively, an elaborately designed approximate binarization function can help to relieve the gradient mismatch in the backward propagation. Bi-Real~\cite{Bi-Real} presents a customized $\mathtt{ApproxSign}$ function  to replace $\mathtt{sign}$ for back-propagation gradient calculation as follow:
\begin{equation}
\mathtt { ApproxSign }(x)=\left\{\begin{array}{ll}{-1,} & {\text {\rm if }\ x<-1} \\ {2 x+x^{2},} & {\text {\rm if }\ -1 \leq x<0} \\ {2 x-x^{2},} & {\text {\rm if }\ 0 \leq x<1} \\ {1,} & {\text {\rm otherwise }}\end{array}\right.
\end{equation}
\begin{equation}
{\frac{\partial \mathtt { ApproxSign }(x)}{\partial x}=\left\{\begin{array}{ll}{2+2 x,} & {\text {\rm if }\ -1 \leq x<0} \\ {2-2 x,} & {\text {\rm if }\ 0 \leq x<1} \\ {0,} & {\text {\rm otherwise }}\end{array}\right.}
\end{equation}
Compared to the traditional STE, $\mathtt{ApproxSign}$ has a close shape to that of the original binarization function $\mathtt{sign}$, and thus the gradient error can be controlled to some extent. Circulant Binary Convolutional Networks (CBCN)~\cite{CirculantBNN} also applied an approximate function to address the gradient mismatch from $\mathtt{sign}$ function. Binary Neural Networks+ (BNN+)~\cite{BNN+} directly proposed an improved approximation to the derivative of the $\mathtt{sign}$ function, and introduced a regularization function that encourages the learned weights around the binary values.

\begin{figure}[htbp]
	\begin{center}
		\includegraphics[width=1\linewidth]{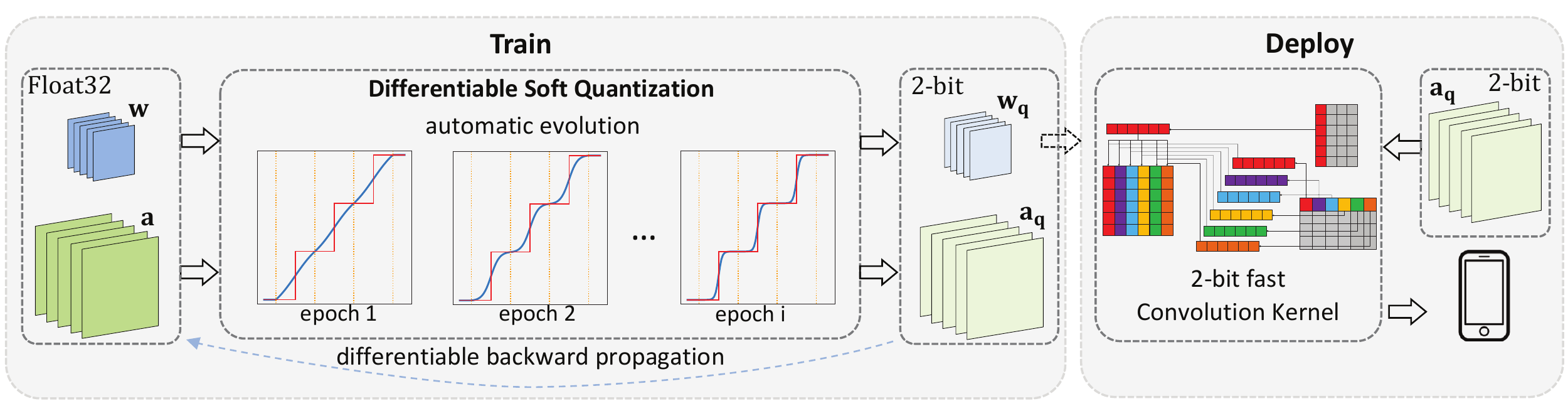}
	\end{center}
	\vspace{-0.2in}
	\caption{Overview of Differentiable Soft Quantization}
	\label{fig:dsq}
	\vspace{-0.1in}
\end{figure}

{Besides focusing on the backward propagation, some recent methods attempted to pursue the good quantization functions in forward propagation, which can also reduce the gradient error.} In~\cite{HWGQ}, the proposed Half-ware Gaussian Quantization (HWGQ) method gave a low-precision estimation for the more commonly used $\mathtt{ReLU}$ function in the forward propagation in training process, which surprisingly works well to solve the gradient mismatch problem. Following the same intuition, Gong \etal present a Differential Soft Quantization (DSQ) method~\cite{Gong:iccv19}, replacing the traditional quantization function with a soft quantization function:
\begin{equation}
\varphi(x)=s \tanh \left(k\left(x-m_{i}\right)\right), \quad \text {\rm if }\ x \in \mathcal{P}_{i}
\end{equation}
where $k$ determines the shape of the asymptotic function, $s$ is a scaling factor to make the soft quantization function smooth and $m_i$ is the center of the interval $\mathcal{P}_{i}$. DSQ can adjust the cutoff value and the shape of the soft quantization function to gradually approach the standard $\mathtt{sign}$ function. {In fact, the DSQ function rectifies the data distribution in a steerable way, and thus helps to alleviate the gradient mismatch.} The overview of DSQ is shown in Figure \ref{fig:dsq}. A similar method~\cite{quantization_networks} also provides a simple and uniform way for weight and activation quantization by formulating it as a differentiable non-linear function. Besides, ProxQuant proposed in~\cite{proxquant} formulates quantized network training as a regularized learning problem instead and optimizes it via the prox-gradient method. ProxQuant does backward propagation on the underlying full-precision vector and applies an efficient prox-operator in between stochastic gradient steps. \cite{selfBN} and~\cite{ImprovedTraining} also explored the smooth transitions for the derivative of the $\mathtt{Sign}$, and used $\mathtt{Tanh}$ function with parameters $v$ and $\mathtt{SoftSign}$ function to reduce training gradient error. {The IR-Net proposed in~\cite{IRNet} included a self-adaptive Error Decay Estimator (EDE) to reduce the gradient error in training, which considers different requirements on different stages of training process and balances the update ability of parameters and reduction of gradient error. The IR-Net provided a new perspective for improving BNNs that retaining both forward and backward information is crucial for accurate BNNs, and it is the first to design BNNs considering both forward and backward information retention.}

Besides modifying binarization function in backward or forward propagation,~\cite{BCGD} directly calibrates the gradients by the blended coarse gradient descent (BCGD) algorithm.
The weight update of BCGD goes by a weighted average of the full-precision weights and their quantized counterparts:
\begin{equation}
\mathbf{w}^{t+1}=(1-\rho) \mathbf{w}^{t}+\rho \mathbf{b}_\mathbf{w}^{t}-\eta \nabla f\left(\mathbf{b}_\mathbf{w}^{t}\right)
\end{equation}
where $\mathbf{w}^t$ denotes the full-precision weights on the $t$-th step and $\nabla f\left(\mathbf{b}_\mathbf{w}^{t}\right)$ denotes the gradient of $\mathbf{b}_\mathbf{w}^t$, thereby yielding sufficient descent in the objective value and thus accelerates the training.~\cite{DeeperUnderstanding} further investigated training methods for quantized neural networks from a theoretical viewpoint, and show that training algorithms that exploit high-precision representations have an important greedy search phase that purely quantized training methods lack, which explains the difficulty of training using low-precision arithmetic.

\subsection{{Efficient Computing Architectures for Binary Neural Networks}}
The most attractive point of binary neural networks is that they enjoy the advantages of fast computation, low power consumption and low memory footprint, which can faithfully support the general hardware (including FPGA, ASIC, CPU, \etc) with limited computational resources. FPGAs are the most widely used platforms because they allow for customizing data paths and adjusting the designs. In particular, FPGAs allow optimization around XNOR-Bitcount operations. ASICs have the potential to provide ultimate power and computational efficiency for binary neural networks, because the hardware layout in ASICs can be designed according to network structure. To make the binarization algorithms more practical in the wide scenarios with different hardware environment, researchers also devoted great efforts to developing hardware-friendly binary networks.

XNOR.AI team, who proposed XNOR-Net~\cite{XNOR-Net}, successfully launched XNOR-Net on the cheap Raspberry Pi device. In order to reduce the amount of computation, they conducted optimization for different targeted hardware. They also tried to combine XNOR-Net with real-time detection algorithms such as YOLO~\cite{yolov3}, and deployed them in the edge computing scenarios like smart home and autonomous driving. FP-BNN~\cite{FP-BNN} implemented a 64-channel acceleration on the Stratix-V FPGA system and analyzed the performance through the Resource-Aware Model Analysis (RAMA) method. Both~\cite{Accelerating-Binarized} and~\cite{FINN} from Xilinx also studied the FPGA-based binary network accelerator using different strategies.~\cite{Accelerating-Binarized} depended on variable-length buffers and achieved up to twice the number of operations per second of existing FPGA accelerators.~\cite{TowardsFastandEnergy} proposed two types of fast and energy-efficient architectures for binary neural network inference. By reusing the results from previous computation, much cycles for data buffer access and computations can be skipped. In order to achieve the most possible memory latency hiding,~\cite{FINN} designed a multi-stream architecture, and applied the Bitcount, Threshhold and OR operations to map the binary network to the FPGA operators. The researchers of the Haas-Platna Software Institute in Germany implemented an accelerated version of BMXNet~\cite{bmxnet,bmxnetv2} on GPU for both binary neural networks and linear quantization networks based on MXNet, supporting XNOR-Net and DoReFa-Net. {For ARM platform, engineers from JD company developed the binarization inference library daBNN~\cite{dabnn} for mobile phone platforms. The library uses ARM assembly instructions, which is 8-24$\times$ more efficient than BMXNet.}

\begin{table}[htb]
\setlength{\abovecaptionskip}{0.cm}

\centering
\tiny
\caption{Deployment Performance of Binary Neural Networks}
\label{BNN-in-different-platforms}
\begin{threeparttable}
\setlength{\tabcolsep}{0.5mm}{
\begin{tabular}{|c|c|c|c|c|c|c|c|c|c|}
\hline
\textbf{Dataset} & \textbf{Method} & \multicolumn{1}{c|}{\textbf{\tabincell{c}{Acc.\\(\%)}}} & \textbf{Topology} & \textbf{Platform} & \multicolumn{1}{c|}{\textbf{LUTs}} & \multicolumn{1}{c|}{\textbf{BRAMs}} & \multicolumn{1}{c|}{\textbf{\tabincell{c}{Clk\\(MHz)}}} & \multicolumn{1}{c|}{\textbf{FPS}} & \multicolumn{1}{c|}{\textbf{\tabincell{c}{Power\\(W)}}} \\ \cline{1-10}
\multirow{6}{*}{\tabincell{c}{MNIST\\\cite{MNIST}}} & FINN-R~\cite{FINN-R} & 97.7 & MLP-4 & Zynq-8020 & 25,358 & 220 & 100 & - & 2.5 \\ \cline{2-10}
 & FINN-R~\cite{FINN-R} & 97.7 & MLP-4 & ZynqUltra 3EG & 38,250 & 417 & 300 & - & 11.8 \\ \cline{2-10}
 & ReBNet\cite{ReBNet} & 98.3 & MLP-4* & Spartan 750 & 32,600 & 120 & 200 & - & -  \\ \cline{2-10}
 & FINN~\cite{FINN} & 98.4 & MLP-4 & Zynq-7045 & 82,988 & 396 & 200 & 1,561,000 & 22.6 \\ \cline{2-10}
 & BinaryEye~\cite{BinaryEye} & 98.4 & MLP-4 & Kintex 7325T & 40,000 & 110 & 100 & 10,000 & 12.2 \\ \cline{1-10}
\multirow{3}{*}{\tabincell{c}{SVHN\\\cite{SVHN}}} & FINN~\cite{FINN} & 94.9 & CNV-6 & Zynq-7045 & 46,253 & 186 & - & 21,900 & 11.7 \\ \cline{2-10}
 & FBNA~\cite{FBNA} & 96.9 & CNV-6 & Zynq-7020 & 29,600 & 103 & - & 6,451 & 3.2 \\ \cline{2-10}
 & ReBNet~\cite{ReBNet} & 97.0 & CNV-6* & Zynq-7020 & 53,200 & 280 & 100 & - & -  \\ \cline{1-10}
\multirow{15}{*}{\tabincell{c}{CIFAR-10\\\cite{LearningMultipleLayers}}} & Zhou \etal~\cite{Zhou2017DeepLB} & 66.6 & CNV-2* & Zynq-7045 & 20,264 & - & - & - & -  \\ \cline{2-10}
 & Nakahara \etal~\cite{A-memory-based-realization} & - & CNV* & Vertex-7 690T & 20,352 & 372 & 450 & - & 15.4 \\ \cline{2-10}
 & Fraser \etal~\cite{Fraser2017ScalingBN} & 79.1 & 1/4 cnn* & KintexUltra 115 & 35,818 & 144 & 125 & 12,000 & -  \\ \cline{2-10}
 & FINN-R~\cite{FINN-R} & 80.1 & CNV-6 & ZynqUltra 3EG & 41,733 & 283 & 300 & - & 10.7 \\ \cline{2-10}
 & FINN-R~\cite{FINN-R} & 80.1 & CNV-6 & Zynq-7020 & 25,700 & 242 & 100 & - & 2.3 \\ \cline{2-10}
 & FINN~\cite{FINN} & 80.1 & CNV-6 & Zynq-7045 & 46,253 & 186 & 200 & 21,900 & 11.7 \\ \cline{2-10}
 & FINN~\cite{FINN-R} & 80.1 & CNV-6 & ADM-PCIE-8K5 & 365,963 & 1,659 & 125 & - & 41.0 \\ \cline{2-10}
 & FINN~\cite{A-fully-connected-layer} & 80.1 & CNV-6 & Zynq-7020 & 42,823 & 270 & 166 & 445 & 2.5 \\ \cline{2-10}
 & Nakahara \etal~\cite{A-fully-connected-layer} & 81.8 & CNV-6 & Zynq-7020 & 14,509 & 32 & 143 & 420 & 2.3 \\ \cline{2-10}
 & Fraser \etal~\cite{Fraser2017ScalingBN} & 85.2 & 1/2 cnn* & KintexUltra 115 & 93,755 & 386 & 125 & 12,000 & -  \\ \cline{2-10}
 & Zhou \etal~\cite{Zhou2017DeepLB} & 86.1 & CNV-5* & Vertex-7 980T & 556,920 & - & 340 & 332,158 & -  \\ \cline{2-10}
 & ReBNet~\cite{ReBNet} & 87.0 & CNV-6* & Zynq-7020 & 53,200 & 280 & 200 & - & -  \\ \cline{2-10}
 &  Zhao \etal~\cite{Accelerating-Binarized} & 87.7 & CNV-6 & Zynq-7020 & 46,900 & 140 & 143 & 168 & 4.7 \\ \cline{2-10}
 & Fraser \etal~\cite{Fraser2017ScalingBN} & 88.3 & BNN* & KintexUltra 115 & 392,947 & 1814 & 125 & 12,000 & -  \\ \cline{2-10}
 & FBNA~\cite{FBNA} & 88.6 & CNV-6 & Zynq-7020 & 29,600 & 103 & - & 520 & 3.3 \\ \cline{1-10}
\multirow{2}{*}{\tabincell{c}{ImageNet\\\cite{Deng2009ImageNet}}} & ReBNet~\cite{ReBNet} & 41.0 & CNV-5* & VertexUltra 095 & 1,075,200 & 3456 & 200 & - & -  \\ \cline{2-10}
 & Yonekawa \etal~\cite{On-Chip-Memory-Based} & - & VGG-16 & ZynqUltra 9EG & 191,784 & 32,870 & 150 & 31.48 & 22.0 \\ \hline
\end{tabular}}
\begin{tablenotes}
    \footnotesize
    \item[1] The * represents the network using a customized network structure, and the Acc. in table refers to Top-1 classification accuracy on each dataset.
\end{tablenotes}
\end{threeparttable}
\end{table}

We provide a comparison of different binary neural network implementations~\cite{FINN-R,FINN,ReBNet,BinaryEye,A-memory-based-realization,Fraser2017ScalingBN,A-fully-connected-layer,Accelerating-Binarized,On-Chip-Memory-Based,Zhou2017DeepLB} on different FPGA platforms in Table \ref{BNN-in-different-platforms}. It can be seen that the method proposed by~\cite{On-Chip-Memory-Based} can achieve comparable accuracy to full-precision models, although it is not efficient enough. The implementation of Xilinx's~\cite{FINN} owns the most promising speed with a low power consumption. A series of experiments prove that it can achieve a good balance among accuracy, speed and power consumption.~\cite{ReBNet} obtains high accuracy on small datasets such as MNIST and CIFAR-10, but a poor result on ImageNet. We have to point out that despite the progress of developing hardware-friendly algorithms, till now there have been quite few binary models that can perform well on large datasets such as ImageNet in terms of both speed and accuracy.

\subsection{Applications of Binary Neural Networks}
Image classification is a fundamental task in computer vision and machine learning. Therefore, most of existing studies chose to evaluate binary neural networks on the image classification tasks. {BNNs can greatly accelerate and compress the neural network models, which is of great attraction to deep learning researchers. Both weights and activations are binary in BNNs and it results in $58\times$ faster convolutional operations and $32\times$ memory savings theoretically. Thus the binary neural networks are also applied to other common tasks such as object detection and semantic segmentation.}

In the literature, Kung \etal utilized binary neural networks for both object recognition and image classification tasks~\cite{Kung2018} on infrared images. In this work, the binary neural networks got comparable performance to full-precision networks on MNIST and IR datasets and achieved at least a 4$\times$ acceleration and an energy saving of three orders of magnitude over the GPU.~\cite{BWBDN} also addressed the fast object detection algorithm by unifying the prediction and object detection process. It obtained 62$\times$ acceleration and saved 32$\times$ storage space using binary VGG-16 network, where except the last convolution layer, all other layers are binarized. Li \etal generated quantized object detection neural networks based on RetinaNet and faster R-CNN, and show that these detectors achieve very encouraging performance~\cite{Li_2019_CVPR}. {Leng \etal applied BNNs to different tasks, and evaluated their method on convolutional neural networks for image classification and object detection, and recurrent neural networks for language model~\cite{Leng2017ExtremelyLB}.} Zhuang \etal proposed a ``network decomposition" strategy called Group-Net, which shows strong generalization to different tasks including classification and semantic segmentation, outperforming the previous best binary neural networks in terms of accuracy and major computation savings~\cite{Zhuang_2019_CVPR}. In~\cite{SeerNet}, SeerNet considers feature-map sparsity through low-bit quantization, which is applicable to general convolutional neural networks and tasks.

{The researchers also studied enhancing the robustness of neural network models through model binarization. Binary models were generally considered to be more robust than full-precision models, because they were considered to filter the noise of input. Lin \etal explored the impact of quantization on the model robustness. They showed that quantization operations on parameters help BNNs to reduce the distance by removing perturbations when magnitude of noise is small. However, for vanilla BNNs, the distance is enlarged when magnitude of noise is large. The inferior robustness comes from the error amplification effect in forward propagation of BNNs, where the quantization operation further enlarges the distance caused by amplified noise. They propose Defensive Quantization (DQ)~\cite{lin2018defensive} to defend the adversarial examples for quantized models by suppressing the noise amplification effect and keeping the magnitude of the noise small in each layer. Quantization improves robustness instead of making it worse in DQ models, thus they are even more robust than full-precision networks.}

\subsection{Tricks for Training Binary Neural Networks}
Due to the highly discrete nature of the binarization, training binary neural networks often requires the introduction of special training techniques to make the training process more stable and the convergence accuracy much higher. In this section, we summarize the general and effective binary neural network training techniques that have been widely adopted in the literature, from the aspects including network structure transformation, optimizer and hyper-parameter selection, gradient approximation and asymptotic quantization.

\subsubsection{Network Structure Transformation}
Binarization converts activations and weights to $\{-1, +1\}$. This is actually equivalent to regularizing the data, making the data distribution changed in an unexpected way after binarization. Adjusting the network structure serves as a promising solution to adapting to the distribution changes.

{Simply reordering the layers in the network can improve the performance of the binary neural network.} In~\cite{alizadeh2018a}, researchers from the University of Oxford pointed out that almost all binarization studies have repositioned the location of the pooling layer. The pooling layer is always used immediately after the convolutional layer to avoid information loss caused by max pooling after binarization. Experiments have shown that this position reorder has a great improvement in accuracy. In addition to the pooling layer, the location of the batch normalization layer also greatly affects the stability of binary neural network training.~\cite{TSQ} and~\cite{HWGQ} insert a batch normalization layer before all quantization operations to rectify the data. After this transformation, the quantized input obeys a stable distribution (sometimes close to Gaussian), and thus the mean and variance keep within a reasonable range and the training process becomes much smoother. 

Based on the similar idea, instead of adding new layers, several recent work attempts to directly modify the network structure. For example, Bi-Real~\cite{Bi-Real} connects the full-precision feature maps across the layer to the subsequent network. This method essentially adjusts the data distribution through structural transformation. Mishra \etal devised Wide Reduced-Precision Networks (WRPN)~\cite{WRPN}, which increase the number of filters in each layer and thus reform the data distributions. Binary Ensemble Neural Network (BENN)~\cite{BENN} leverages the ensemble method to fit the underlying data distributions. Liu \etal proposed circulant filters (CiFs) and a circulant binary convolution (CBConv) to enhance the capacity of binarized convolutional features, and circulant back propagation (CBP) was also proposed to train the structures~\cite{CirculantBNN}. {BBG~\cite{DBLP:journals/corr/abs-1909-12117} even appended a gated residual to compensate their information loss during the forward process.}

\subsubsection{Optimizer and Hyper-parameter Selection}
{Choosing the proper hyper-parameters and specific optimizers when training binary neural networks also improves the performance of BNNs.} Most existing binary neural network models chose an adaptive learning rate optimizer, such as Adam. Using Adam can make the training process better and faster, and the smoothing coefficient of the second derivative is especially critical. The analysis by~\cite{alizadeh2018a} shows that if using a fixed learning rate optimizer that does not consider historical information, such as a stochastic gradient descent (SGD) algorithm, one needs to adopt a large batch size to improve the performance.

The setting of the batch normalization's momentum coefficient is also critical. In~\cite{alizadeh2018a}, by comparing the precision results under different momentum coefficients, it is found that the parameters of the batch normalization need to be set appropriately to adapt to the jitter caused by the binarization operation.

\subsubsection{Asymptotic Quantization}
{Since the quantization has negative impact on training, many methods employed the asymptotic quantization strategy, which gradually increases the degree of quantization, to reduce the losses caused by parameter binarization.} Practice shows that this step-by-step quantization method is useful to find the optimal solution. For instance, INQ~\cite{INQ} groups the parameters and gradually increases the number of groups participating in the quantization to achieve group-based step-by-step quantization. {\cite{Zhuang_2018_CVPR} introduces the idea of stepping the bit-width, which first quantizes to a higher bit-width and then quantizes to a lower bit-width.} This strategy can help to avoid the large perturbations caused by extremely low-bit quantization, compensating the gradient error of quantized parameters during training.

\subsubsection{Gradient Approximation}
{It became a common practice to use a smoother estimator in binary neural network training process.} The gradient error usually exists in backward propagation due to the straight-through estimator. Finding an approximate function close to the binarization function serves as the simple and practical solution. {This becomes a popular technique widely considered in recent studies~\cite{Bi-Real,HWGQ,CirculantBNN,BNN+,selfBN,ImprovedTraining,IRNet}, where the approximate functions are tailored according to different motivations, to replace the standard $\mathtt{clip}$ function that causes gradient error.} For designing a proper approximate function, an inspiring idea is to align its shape with that of the binarization function~\cite{Gong:iccv19}.

\section{Evaluation and Discussions}
\label{section4}

\subsection{Datasets and Network Structures}
To evaluate the binary neural network algorithms, the image classification task is widely chosen, and correspondingly two common image datasets: CIFAR-10~\cite{LearningMultipleLayers} and ImageNet~\cite{Deng2009ImageNet} are usually used. The CIFAR-10 is a relatively small dataset containing 60,000 images with 10 categories, while ImageNet dataset is currently the most popular image classification dataset. For other tasks like object detection and semantic segmentation, PASCAL VOC~\cite{Everingham:2010:PVO:1747084.1747104} and COCO (Common Objects in Context)~\cite{DBLP:journals/corr/LinMBHPRDZ14} are also employed for evaluating the performance of the binary neural networks. {PASCAL VOC dataset is derived from the PASCAL Visual Object Classes challenge, which is used to evaluate the performance of models for various tasks in the field of computer vision.} Many excellent computer vision models (including classification, positioning, detection, segmentation, motion recognition, \etc) are based on the PASCAL VOC dataset, especially some object detection models. COCO is a dataset provided by the Microsoft team for image recognition and object detection. It collects images by searching 80 object categories and various scene types such as Flickr.

To investigate the generalization capability of the binary neural network algorithm over different network structures, various deep models including VGG~\cite{simonyan2015very}, AlexNet~\cite{krizhevsky2012imagenet}, ResNet-18~\cite{he2016deep}, ResNet-20, ResNet-34, and ResNet-50, \etc. will be binarized and tested. These models have outstanding contributions in the progress of deep learning, and make significant breakthrough in ImageNet classification task. Among them, the VGG network contains a large number of parameters and convolution operations, so binarizing VGG can obviously show the inference acceleration of different algorithms. ResNet is currently the most popular deep model in many tasks, with a sufficient number of layers.

\subsection{Image Classification Tasks}

\begin{table}[!h]
\setlength{\abovecaptionskip}{0.cm}

\caption{{Image Classification Performance of Binary Neural Networks on CIFAR-10 Dataset}}
\label{BNN-On-CIFAR-10}
\centering
\scriptsize

\setlength{\tabcolsep}{1.8mm}{
\begin{tabular}{|c|c|c|c|c|c|c|}
\hline
\multicolumn{2}{|c|}{\textbf{Type}}  & \multicolumn{1}{c|}{\textbf{Method}} & \multicolumn{1}{c|}{\textbf{\begin{tabular}[c]{@{}c@{}}Bit-Width\\(W/A)\end{tabular}}} & \multicolumn{1}{c|}{\textbf{Topology}} & \multicolumn{1}{c|}{\textbf{\begin{tabular}[c]{@{}c@{}}Acc.\\ (\%)\end{tabular}}} \\ \hline

\multicolumn{2}{|c|}{\multirow{6}{*}{\tabincell{c}{Full-Precision\\Neural Networks}}} & \multirow{6}{*}{-} & 32/32 & VGG-Small~\cite{LQ-Net} & 93.8 \\ \cline{4-6}

\multicolumn{2}{|c|}{} &   & 32/32 & ResNet-20~\cite{LQ-Net} & 92.1 \\ \cline{4-6}

\multicolumn{2}{|c|}{} & &  32/32 & ResNet-32~\cite{proxquant} & 92.8 \\ \cline{4-6}

\multicolumn{2}{|c|}{} & &  32/32 & ResNet-44~\cite{proxquant} & 93.0 \\ \cline{4-6}

\multicolumn{2}{|c|}{} & &  32/32 & VGG-11~\cite{Subsidiary} & 83.8 \\ \cline{4-6}

\multicolumn{2}{|c|}{} & &  32/32 & NIN~\cite{Subsidiary} & 84.2 \\ \cline{1-6}

\multicolumn{2}{|c|}{\multirow{2}{*}{\tabincell{c}{Naive Binary\\ Neural Networks}}} & BinaryConnect~\cite{BinaryConnect} & 1/32 & VGG-Small & 91.7 \\ \cline{3-6} 

\multicolumn{2}{|c|}{} & BNN~\cite{BNN} &  1/1 & VGG-Small & 89.9 \\ \hline

\multirow{31}{*}{\tabincell{c}{{Optimization}\\\\{Based}\\\\{Binary}\\\\{Neural}\\\\{Networks}}}& \multirow{12}{*}{\tabincell{c}{Minimize\\the\\Quantization\\Error}} & BWN~\cite{XNOR-Net} & 1/32 & VGG-Small & 90.1 \\ \cline{3-6}

&  & \multirow{2}{*}{XNOR-Net~\cite{XNOR-Net}} &  1/1 & VGG-Small & 89.8 \\ \cline{4-6}

&  & & 1/1 & Customized~\cite{HORQ} & 77.0 \\ \cline{3-6}

&  & \multirow{2}{*}{DoReFa-Net~\cite{DoReFa-Net}} & 1/32 & ResNet-20 & 90.0 \\ \cline{4-6}

&  & &  1/1 & ResNet-20 & 79.3 \\ \cline{3-6}

&  & HORQ~\cite{HORQ} & 2/1 & Customized~\cite{HORQ} & 82.0 \\ \cline{3-6}

&  & TSQ~\cite{TSQ} & 3/2 & VGG-Small & 93.5 \\ \cline{3-6}

&  & \multirow{3}{*}{{BBG~\cite{DBLP:journals/corr/abs-1909-12117}}} & {1/1} & {ResNet-20} & {85.3} \\ \cline{4-6}

&  & &  {1/1} & {ResNet-20 (2x)} & {90.7} \\ \cline{4-6}

&  & &  {1/1} & {ResNet-20 (4x)} & {92.5} \\ \cline{3-6}

&  & \multirow{2}{*}{LQ-Nets~\cite{LQ-Net}} & 1/32 & ResNet-20 & 90.1 \\ \cline{4-6}

&  & & 1/2 & VGG-Small & 93.4 \\ \cline{2-6}

&\multirow{13}{*}{\tabincell{c}{Improve\\Network\\Loss\\Function}} & \multirow{2}{*}{LAB~\cite{Loss-Aware-BNN}} & 1/32 & VGG-Small & 89.5 \\ \cline{4-6}

&  & &  1/1 & VGG-Small & 87.7 \\ \cline{3-6} 

&  & \multirow{3}{*}{\tabincell{c}{Main/Subsidiary\\Network~\cite{Subsidiary}}} & 1/1 & NIN & 83.1 \\ \cline{4-6}

&  & &  1/1 & VGG-11 & 82.0 \\ \cline{4-6} 

&  & &  1/1 & ResNet-18 & 86.4 \\ \cline{3-6}
 
&  & \multirow{2}{*}{BCGD~\cite{Subsidiary}} & 1/4 & VGG-11 & 89.6 \\ \cline{4-6}

&  & &  1/4 & ResNet-20 & 90.1 \\ \cline{3-6}
  
&  & \multirow{3}{*}{ProxQuant~\cite{proxquant}} & 1/32 & ResNet-20 & 90.7 \\ \cline{4-6}

&  & &  1/32 & ResNet-32 & 91.5 \\ \cline{4-6}

&  & &  1/32 & ResNet-44 & 92.2 \\ \cline{3-6}
  
&  & BNN-DL~\cite{Regularize-act-distribution} &  1/1 & VGG-Small & 90.0 \\ \cline{3-6}

&  & \multirow{2}{*}{{CI-BCNN~\cite{LearningChannel-Wise}}} & {1/1} & {VGG-Small} & {92.5} \\ \cline{4-6}

&  & &  {1/1} & {ResNet-20} & {91.1} \\ \cline{2-6}

& \multirow{7}{*}{\tabincell{c}{Reduce the\\Gradient Error}} & \multirow{3}{*}{\tabincell{c}{DSQ~\cite{Gong:iccv19}}} &  \tabincell{c}{1/1} & VGG-Small & 91.7 \\ \cline{4-6}

&  & &  \tabincell{c}{1/32} & ResNet-20 & 90.2 \\ \cline{4-6}

&  & &  \tabincell{c}{1/1} & ResNet-20 & 84.1 \\ \cline{3-6}

&  & \multirow{4}{*}{\tabincell{c}{{IR-Net~\cite{Gong:iccv19}}}} &  \tabincell{c}{{1/32}} & {ResNet-20} & {90.2} \\ \cline{4-6}

&  & &  \tabincell{c}{{1/1}} & {VGG-Small} & {90.4} \\ \cline{4-6}

&  & &  \tabincell{c}{{1/1}} & {ResNet-18} & {91.5} \\ \cline{4-6}

&  & &  \tabincell{c}{{1/1}} & {ResNet-20} & {86.5} \\ \hline

\end{tabular}}
\end{table}

\begin{table}[]
\setlength{\abovecaptionskip}{0.cm}
\caption{{Image Classification Performance of Binary Neural Networks on ImageNet Dataset}}
\label{BNN-On-ImageNet}
\centering
\scriptsize
\setlength{\tabcolsep}{0.6mm}{
\begin{tabular}{|c|c|c|c|c|c|c|c|}
\hline
\multicolumn{2}{|c|}{\textbf{Type}} & \multicolumn{1}{c|}{\textbf{Method}} & \multicolumn{1}{c|}{\textbf{\begin{tabular}[c]{@{}c@{}}Bit-Width\\(W/A)\end{tabular}}} & \textbf{Topology} & \multicolumn{1}{c|}{\textbf{\begin{tabular}[c]{@{}c@{}}Top-1\\(\%)\end{tabular}}} & \multicolumn{1}{c|}{\textbf{\begin{tabular}[c]{@{}c@{}}Top-5\\(\%)\end{tabular}}} \\ \hline

\multicolumn{2}{|c|}{\multirow{5}{*}{\tabincell{c}{Full-Precision\\Neural Networks}}} & \multirow{5}{*}{\tabincell{c}{-}} & 32/32 & AlexNet~\cite{LQ-Net} & 57.1 & 80.2 \\ \cline{4-7}

\multicolumn{2}{|c|}{} &  & 32/32 & ResNet-18~\cite{LQ-Net} & 69.6 & 89.2 \\ \cline{4-7}

\multicolumn{2}{|c|}{} &  & 32/32 & ResNet-34~\cite{LQ-Net} & 73.3 & 91.3 \\ \cline{4-7}

\multicolumn{2}{|c|}{} &  & 32/32 & ResNet-50~\cite{LQ-Net} & 76.0 & 93.0 \\ \cline{4-7}

\multicolumn{2}{|c|}{} &  & 32/32 & VGG-Variant~\cite{LQ-Net} & 72.0 & 90.5 \\ \cline{1-7}

\multicolumn{2}{|c|}{\multirow{2}{*}{\tabincell{c}{Naive Binary\\Neural Networks}}} & \tabincell{c}{\tabincell{c}{BinaryConnect~\cite{BinaryConnect}}} & 1/32 & AlexNet & 35.4 & 61.0 \\ 
\cline{3-7}

\multicolumn{2}{|c|}{} & \tabincell{c}{BNN~\cite{BNN}} & 1/1 & AlexNet & 27.9 & 50.4 \\ \hline

\multirow{34}{*}{\tabincell{c}{{Optimization}\\\\{Based}\\\\{Binary}\\\\{Neural}\\\\{Networks}}}& \multirow{26}{*}{\tabincell{c}{Minimize\\the\\Quantization\\Error}} & \multirow{2}{*}{BWN~\cite{XNOR-Net}} & 1/32 & AlexNet & 56.8 & 79.4 \\ \cline{4-7}

& & & 1/32 & ResNet-18 & 60.8 & 83.0 \\ \cline{3-7}

& & XNOR-Net~\cite{XNOR-Net} & 1/1 & AlexNet & 44.2 & 69.2 \\ \cline{3-7}

& & \multirow{2}{*}{DoReFa-Net~\cite{DoReFa-Net}} & 1/1 & AlexNet & 43.6 & - \\ \cline{4-7}

& & & 1/1 & AlexNet & 49.8 & - \\ \cline{3-7}

& & \multirow{4}{*}{ABC-Net~\cite{ABC-Net}} & 1/32 & ResNet-18 & 62.8 & 84.4 \\ \cline{4-7}

& & & 2/32 & ResNet-18 & 63.7 & 85.2 \\ \cline{4-7}

& & & 1/1 & ResNet-18 & 42.7 & 67.6 \\ \cline{4-7}

& & & 1/1 & ResNet-34 & 52.4 & 76.5 \\ \cline{3-7}

& & TSQ~\cite{TSQ} & 1/1 & AlexNet & 58.0 & 80.5 \\ \cline{3-7}

& & \multirow{2}{*}{BWNH~\cite{BWN-Hashing}} & 1/32 & AlexNet & 58.5 & 80.9 \\ \cline{4-7}

& & & 1/32 & ResNet-18 & 64.3 & 85.9 \\ \cline{3-7}

& & \multirow{3}{*}{PACT~\cite{PACT}} & 1/32 & ResNet-18 & 65.8 & 86.7 \\ \cline{4-7}

& & & 1/2 & ResNet-18 & 62.9 & 84.7 \\ \cline{4-7}

& & & 1/2 & ResNet-50 & 67.8 & 87.9 \\ \cline{3-7}

& & \multirow{5}{*}{LQ-Nets~\cite{LQ-Net}} & 1/2 & ResNet-18 & 62.6 & 84.3 \\ \cline{4-7}

& & & 1/2 & ResNet-34 & 66.6 & 86.9 \\ \cline{4-7}

& & & 1/2 & ResNet-50 & 68.7 & 88.4 \\ \cline{4-7}

& & & 1/2 & AlexNet & 55.7 & 78.8 \\ \cline{4-7}

& & & 1/2 & VGG-Variant & 67.1 & 87.6 \\ \cline{3-7}

& & \multirow{3}{*}{SYQ~\cite{SYQ}} & 1/2 & AlexNet & 55.4 & 78.6 \\ \cline{4-7}

& & & 1/8 & ResNet-18 & 62.9 & 84.6 \\ \cline{4-7}

& & & 1/8 & ResNet-50 & 70.6 & 89.6 \\ \cline{3-7}

& & \multirow{3}{*}{WRPN~\cite{WRPN}} & \begin{tabular}[c]{@{}r@{}}1/1 (1$\times$)\end{tabular} & ResNet-34 & 60.5 & - \\ \cline{4-7}

& & & \begin{tabular}[c]{@{}r@{}}1/1 (2$\times$)\end{tabular} & ResNet-34 & 69.9 & - \\ \cline{4-7}

& & & \begin{tabular}[c]{@{}r@{}}1/1 (3$\times$)\end{tabular} & ResNet-34 & 72.4 & - \\ \cline{3-7}

& & \multirow{2}{*}{{XNOR-Net++~\cite{Bulat2019XNORNetIB}}} & \begin{tabular}[c]{@{}r@{}}{1/1 (1$\times$)}\end{tabular} & {ResNet-18} & {57.1} & {79.9} \\ \cline{4-7}

& &  & \begin{tabular}[c]{@{}r@{}}{1/1 (1$\times$)}\end{tabular} & {AlexNet} & {46.9} & {71.0} \\ \cline{2-7}

& \multirow{9}{*}{\tabincell{c}{Improve\\Network\\Loss\\Function}} & INQ~\cite{INQ} & 2/32 & ResNet-18 & 66.0 & 87.1 \\ \cline{3-7}

& & \tabincell{c}{BNN-DL~\cite{Regularize-act-distribution}} & 1/1 & AlexNet & 41.3 & 65.8 \\ \cline{3-7}

& & \tabincell{c}{XNOR-Net-DL~\cite{Regularize-act-distribution}} & 1/1 & AlexNet & 47.8 & 71.5 \\ \cline{3-7}

& & \tabincell{c}{DoReFa-Net-DL~\cite{Regularize-act-distribution}} & 1/1 & AlexNet & 47.8 & 71.5 \\ \cline{3-7}

& & \tabincell{c}{CompactNet-DL~\cite{Regularize-act-distribution}} & 1/2 & AlexNet & 47.6 & 71.9 \\ \cline{3-7}

& & \tabincell{c}{WRPN-DL~\cite{Regularize-act-distribution}} & 1/1 & AlexNet & 53.8 & 77.0 \\ \cline{3-7}

&  &\multirow{2}{*}{\tabincell{c}{Main/Subsidiary\\Network~\cite{Subsidiary}}}& \multirow{2}{*}{1/1} & \multirow{2}{*}{ResNet-18} & \multirow{2}{*}{50.1} & \multirow{2}{*}{-} \\  
& &  &  &  &  & \\ \cline{1-7}
\end{tabular}}
\end{table}

\begin{table}[]\ContinuedFloat
\vspace{-2cm}
\setlength{\abovecaptionskip}{0.cm}
\caption{ ($Cont.$) {Image Classification Performance of Binary Neural Networks on ImageNet Dataset}}
\centering
\scriptsize
\setlength{\tabcolsep}{2mm}{
\begin{tabular}{|c|c|c|c|c|c|c|}
\hline

\multicolumn{2}{|c|}{\textbf{Type}} & \multicolumn{1}{c|}{\textbf{Method}} & \multicolumn{1}{c|}{\textbf{\begin{tabular}[c]{@{}c@{}}Bit-Width\\(W/A)\end{tabular}}} & \textbf{Topology} & \multicolumn{1}{c|}{\textbf{\begin{tabular}[c]{@{}c@{}}Top-1\\(\%)\end{tabular}}} & \multicolumn{1}{c|}{\textbf{\begin{tabular}[c]{@{}c@{}}Top-5\\(\%)\end{tabular}}} \\ \hline

\multirow{18}{*}{\tabincell{c}{{Optimization}\\\\{Based}\\\\{Binary}\\\\{Neural}\\\\{Networks}}}&  & \multirow{2}{*}{\tabincell{c}{{{CI-BCNN~\cite{LearningChannel-Wise}}}}} & {1/1} & {ResNet-18} & {59.9} & {84.2} \\ \cline{4-7}

&  & & {1/1} & {ResNet-34} & {54.9} & {86.6} \\ \cline{2-7}
 &\multirow{16}{*}{\tabincell{c}{ Reduce\\the\\Gradient\\Error}} &\multirow{2}{*}{Bi-Real~\cite{Bi-Real}} & 1/1 & ResNet-18 & 56.4 & 79.5 \\ \cline{4-7}

& & & 1/1 & ResNet-34 & 62.2 & 83.9 \\ \cline{3-7}

& & HWGQ~\cite{HWGQ} & 1/1 & AlexNet & 52.7 & 76.3 \\ \cline{3-7}

& & CBCN~\cite{CirculantBNN} & 1/1 & ResNet-18 & 61.4 & 82.8 \\ \cline{3-7}

& &\multirow{2}{*}{\tabincell{c}{{Quantization}}} & 1/32 & AlexNet & 58.8 & 81.7 \\ \cline{4-7}

& &\multirow{2}{*}{Networks~\cite{quantization_networks}} & 1/32 & ResNet-18 & 66.5 & 87.3 \\ \cline{4-7}

& & & 1/32 & ResNet-50 & 72.8 & 91.3 \\ \cline{3-7}

& & \multirow{2}{*}{BCGD~\cite{BCGD}} & 1/4 & ResNet-18 & 65.5 & 86.4 \\ \cline{4-7}

& & & 1/4 & ResNet-34 & 68.4 & 88.3 \\ \cline{3-7}

& & DSQ~\cite{Gong:iccv19} & 1/32 & ResNet-18 & 63.7 & - \\ \cline{3-7}

& & \multirow{4}{*}{{IR-Net~\cite{IRNet}}} & {1/32} & {ResNet-18} & {62.9} & {84.1} \\ \cline{4-7}
& & & {1/32} & {ResNet-34} & {70.4} & {89.5} \\ \cline{4-7}
& & & {1/1} & {ResNet-18} & {58.1} & {80.0} \\ \cline{4-7}
& & & {1/1} & {ResNet-34} & {66.5} & {86.8} \\ \cline{3-7}

& & \multirow{2}{*}{{IT-BNN~\cite{ImprovedTraining}}} & {1/1} & {ResNet-18} & {53.7} & {76.8} \\ \cline{4-7}
& & & {1/1} & {AlexNet} & {48.6} & {72.8} \\ \cline{4-7}

\hline
\end{tabular}}
\end{table}

{Most binary neural networks adopt the inference accuracy of image classification as the evaluation metric, as the classical classification models do.} Table \ref{BNN-On-CIFAR-10} and \ref{BNN-On-ImageNet}  respectively illustrate the performance of the typical binary neural network methods on CIFAR-10 and ImageNet, and compare the inference accuracy with different bit-width and network structures.

Comparing the performance of binary neural networks on different datasets, we can first observe that binary neural networks can approach the performance of full-precision neural networks on small datasets (\eg MNIST, CIFAR-10), but still suffer a severe performance drop on large datasets (\eg ImageNet). This is mainly because for the large dataset, the binarized network lacks sufficient capacity to capture the large variations among data. This fact indicates that there still require great efforts for pursuing the delicate binarization and optimization solution to design a satisfactory binary neural network.

From the table \ref{BNN-On-CIFAR-10} and \ref{BNN-On-ImageNet}, it can be concluded that the neural networks are more sensitive to the binarization of activations. When only quantizing weights to 1-bit and leaving the activations as full-precision, there is a smaller performance degradation. Taking ResNet-18 in ABC-Net~\cite{ABC-Net} on ImageNet dataset as an example, there is only about 7\% accuracy loss after applying binarization to weights but there is addition 20\% loss after the activations are binarized. Thus eliminating the influence of activation binarization is usually much more important when designing binary network, which becomes the main motivations for studies like~\cite{Regularize-act-distribution} and~\cite{PACT}. After adding reasonable regularization to the distribution of activations, the harmful effect caused by binarization on activations will be reduced, and subsequently the accuracy is naturally improved.

What's more, the robustness of binary neural networks is highly relevant to their structures. Some specific structure patterns are friendly to binarization, such as skip connections proposed in~\cite{Bi-Real} and wider blocks proposed in~\cite{WRPN}. With a shortcut to directly pass full-precision values to the following layers, Bi-Real~\cite{Bi-Real} achieves performance close to full-precision models. With a $3\times$ wider structure, the accuracy loss of ResNet-34 in~\cite{WRPN} is lower than 1\%. In fact, what they essentially do is to enable the information to pass through the whole network as much as possible. Although the structure modification may increase the amount of calculation, they can still get a significant acceleration benefiting from the XNOR-Bitcount operation.

{Different optimization-based methods represent different understandings of BNNs. Among the papers aiming to minimizing the quantization error, many methods that directly reduce the quantization error were proposed to make the binary neural networks approximate full-precision neural networks. These papers believed that the closer the binary parameters are to full-precision parameters, the better the BNNs perform. Another idea is improving the loss function. This type of methods makes the parameter distribution in BNNs friendly to the binarization operation by modifying loss function. Moreover, STE proposed in BinaryConnect is rough, which results in some problems such as gradient mismatch. Thus many recent works use smooth transition such as $\mathtt{Tanh}$ function to reduce the gradient loss, and it became a common practice to use a smoother estimator.} 

{We believe binary neural networks should not be simply regarded as the approximations of full-precision neural networks, more specific designs for the special characteristics of BNNs are necessary. In fact, some of the recent works essentially worked on this such as XNOR-Net++~\cite{Bulat2019XNORNetIB}, CBCN~\cite{CirculantBNN}, Self-Binarizing Networks~\cite{selfBN}, BENN~\cite{BENN}, \etc. The results show that specially designed methods considering characteristics of BNNs can achieve better performance. They prove the view that BNNs need different optimization compared with the full-precision models although they share the same network architecture.}

{It is also worth mentioning that accuracy is not the only criterion of BNNs, the versatility is another key to measure whether a method can be used in practice. Some methods proposed in existing papers are very versatile, such as scale factors proposed in XNOR-Net~\cite{XNOR-Net}, smooth transition~\cite{BNN+}, addition shortcuts~\cite{Bi-Real}, \etc. The methods are versatile because of their simple implementation and low coupling. Thus they become common practices to improve the performance of BNNs. Some methods improve the performance of binary neural networks by designing or learning delicate quantizers. Such quantizers usually have stronger ability to preserve the information. However, we have to point out that some of them suffer complicate computation and even multi-stage training pipelines, which is sometimes unfriendly to hardware implementation and reproducibility. This means it is hard to acquire an effective speed up with such quantizers in real-world deployment. Therefore, purely pursuing high accuracy without considering the acceleration implementation makes no sense in practice. The balance between accuracy and speed is also an essential criterion for binarization research that should be always kept in mind.}

\subsection{Other Tasks}
It is worth noting that most of the current binary neural networks that focus on image classification tasks cannot be directly generalized to other tasks. For different tasks, it is still highly required to design specific binary neural networks for the desirable performance. In addition to image classification task, there are also a few studies that designed and evaluated the binary neural network models for other tasks, such as object detection and semantic segmentation tasks. For the object detection task, Table \ref{BNN-On-COCO} and Table \ref{BNN-On-PASCAL2007} respectively list the performance of different binary neural networks on the COCO 2017 and PASCAL VOC 2007 datasets. For the semantic segmentation tasks, Table \ref{BNN-On-PASCAL2012} compares different binary neural networks on the PASCAL VOC 2012 dataset. The experiments are based on different bit-width and network structures.

\begin{table}[t]
\vspace{-2cm}
\setlength{\abovecaptionskip}{0.cm}
\centering
\caption{{Object Detection Performance of Binary Neural Networks on COCO 2017 Dataset}}
\label{BNN-On-COCO}
\scriptsize
\begin{tabular}{|c|c|c|c|}
\hline
\multicolumn{1}{|c|}{\textbf{Topology}} & \multicolumn{1}{c|}{\textbf{Method}} & \multicolumn{1}{c|}{\textbf{Bit-Width (W/A)}} & \multicolumn{1}{c|}{\textbf{mAP (\%)}} \\ \hline
\multirow{2}{*}{\tabincell{c}{{Faster-RCNN~\cite{Faster-RCNN}}\\{ResNet-18}}} & \multirow{2}{*}{{FQN~\cite{Li_2019_CVPR}}} & \multirow{2}{*}{{1/1}} & \multirow{2}{*}{{28.1}} \\
&&& \\ \cline{1-4}
\multirow{2}{*}{\tabincell{c}{{Faster-RCNN~\cite{Faster-RCNN}}\\{ResNet-34}}} & \multirow{2}{*}{{FQN~\cite{Li_2019_CVPR}}} & \multirow{2}{*}{{1/1}} & \multirow{2}{*}{{31.8}} \\
&&& \\ \cline{1-4}
\multirow{2}{*}{\tabincell{c}{{Faster-RCNN~\cite{Faster-RCNN}}\\{ResNet-50}}} & \multirow{2}{*}{{FQN~\cite{Li_2019_CVPR}}} & \multirow{2}{*}{{1/1}} & \multirow{2}{*}{{33.1}} \\
&&& \\ \cline{1-4}
\multirow{7}{*}{\tabincell{c}{{RetinaNet~\cite{Lin2017Focal}}\\{ResNet-18}}} & Quant whitepaper~\cite{DBLP:journals/corr/abs-1806-08342} & 8/8 & 22.6 \\ \cline{2-4}
 & Integer-only~\cite{DBLP:journals/corr/abs-1712-05877} & 8/8 & 19.7 \\ \cline{2-4}
 & DoReFa-Net~\cite{DoReFa-Net} & 1/1 & 3.9 \\ \cline{2-4}
 & {XNOR-Net~\cite{XNOR-Net}} & {4/4} & {24.4} \\ \cline{2-4}
 & {XNOR-Net (Percentile)~\cite{Li_2019_CVPR}} & {4/4} & {26.7} \\ \cline{2-4}
 & FQN~\cite{Li_2019_CVPR} & 4/4 & 28.6 \\ \hline
\end{tabular}
\vspace{-0.1cm}
\end{table}

\begin{table}[t]
\setlength{\abovecaptionskip}{0.cm}
\centering
\caption{{Object Detection Performance of Binary Neural Networks on PASCAL VOC 2007 Dataset}}
\label{BNN-On-PASCAL2007}
\scriptsize
\setlength{\tabcolsep}{4mm}{
\begin{tabular}{|c|c|c|c|}
\hline
\multicolumn{1}{|c|}{\textbf{Topology}} & \multicolumn{1}{c|}{\textbf{Method}} & \multicolumn{1}{c|}{\textbf{\tabincell{c}{Bit-Width (W/A)}}} & \multicolumn{1}{c|}{\textbf{\tabincell{c}{mAP (\%)}}} \\ \hline
\multirow{3}{*}{\tabincell{c}{{Faster-RCNN~\cite{Faster-RCNN}}\\{VGG}}} & Full-Precision & 32/32 & 68.9 \\ \cline{2-4}
 & BWN~\cite{BWBDN} & 1/32 & 62.5 \\ \cline{2-4}
 & BNN~\cite{BWBDN} & 1/1 & 47.3 \\ \hline
\multirow{3}{*}{\tabincell{c}{{Faster-RCNN~\cite{Faster-RCNN}}\\{AlexNet}}} & Full-Precision & 32/32 & 66.0 \\ \cline{2-4}
 & BWN~\cite{BWBDN} & 1/32 & 62.1 \\ \cline{2-4}
 & BNN~\cite{BWBDN} & 1/1 & 46.4 \\ \hline
 \multirow{2}{*}{\tabincell{c}{{SSD~\cite{Liu2015SSD}}\\{DarkNet}}} & {Full-Precision} & {1/32} & {62.1} \\ \cline{2-4}
 & {ELBNN~\cite{Leng2017ExtremelyLB}} & {2/2} & {62.4} \\ \hline
 \multirow{2}{*}{\tabincell{c}{{SSD~\cite{Liu2015SSD}}\\{VGG-16}}} & {Full-Precision} & {32/32} & {62.1} \\ \cline{2-4}
 & {ELBNN~\cite{Leng2017ExtremelyLB}} & {2/2} & {46.4} \\ \hline
\end{tabular}}
\end{table}

From Table \ref{BNN-On-COCO} and \ref{BNN-On-PASCAL2007} we can see that existing binarization algorithms have achieved encouraging progress for the object detection task, and meanwhile bring the significant acceleration when deployed in real-world systems. But it also should be noted that the binary models still face a great challenge, especially when the activations are quantized to 1-bit. For semantic segmentation task, as shown in Table \ref{BNN-On-PASCAL2012}, the very recent method~\cite{Zhuang_2019_CVPR} achieved high accuracy using only 1-bit, which is almost the same as the full-precision model. But it is unknown how it works and the actual speed up of that method still needs to be verified. 

{Among these results, we found that although the binary neural networks perform well on the classification task, there are still unacceptable losses on other tasks. This makes binary neural networks designed for classification tasks hard to be directly applied to other tasks such as object detection and semantic segmentation. In the classification task, the network pays more attention to the global features, while ignoring the loss of local features caused by binarization. However, local features are more important in other tasks. So when designing binary neural networks for other tasks, the local features of the feature map need to be paid more attention.}

\begin{table}[t]
\vspace{-2cm}
\setlength{\abovecaptionskip}{-0.cm}
\caption{{Semantic Segmentation Performance of Binary Neural Networks on PASCAL VOC 2012 Dataset}}
\label{BNN-On-PASCAL2012}
\centering
\scriptsize
\begin{tabular}{|c|c|c|c|}
\hline
\textbf{Topology} & \multicolumn{1}{c|}{\textbf{Method}} & \textbf{\tabincell{c}{Bit-Width (W/A)}} & \textbf{\tabincell{c}{mAP (\%)}} \\ \hline
\multirow{5}{*}{\tabincell{c}{{Faster-RCNN}\\{ResNet-18}\\{FCN-32s~\cite{Zhuang_2019_CVPR}}}} & Full-precision~\cite{Zhuang_2019_CVPR} & 32/32 & 64.9 \\ \cline{2-4}
 & LQ-Nets~\cite{LQ-Net} & 3/3 & 62.5 \\ \cline{2-4}
 & Group-Net~\cite{Zhuang_2019_CVPR} & 1/1 & 60.5 \\ \cline{2-4}
 & Group-Net + BPAC~\cite{Zhuang_2019_CVPR} & 1/1 & 63.8 \\ \cline{2-4}
 & Group-Net**+BPAC~\cite{Zhuang_2019_CVPR} & 1/1 & 65.1 \\ \hline
\multirow{5}{*}{\tabincell{c}{{Faster-RCNN}\\{ResNet-18}\\{FCN-16s}}} & Full-precision~\cite{Zhuang_2019_CVPR} & 32/32 & 67.3 \\ \cline{2-4}
 & LQ-Nets~\cite{LQ-Net} & 3/3 & 65.1 \\ \cline{2-4}
 & Group-Net~\cite{Zhuang_2019_CVPR} & 1/1 & 62.7 \\ \cline{2-4}
 & Group-Net+BPAC~\cite{Zhuang_2019_CVPR} & 1/1 & 66.3 \\ \cline{2-4}
 & Group-Net**+BPAC~\cite{Zhuang_2019_CVPR} & 1/1 & 67.7 \\ \hline
\multirow{5}{*}{\tabincell{c}{{Faster-RCNN}\\{ResNet-34}\\{FCN-32s}}} & Full-precision~\cite{Zhuang_2019_CVPR} & 32/32 & 72.7 \\ \cline{2-4}
 & LQ-Nets~\cite{LQ-Net} & 3/3 & 70.4 \\ \cline{2-4}
 & Group-Net~\cite{Zhuang_2019_CVPR} & 1/1 & 68.2 \\ \cline{2-4}
 & Group-Net+BPAC~\cite{Zhuang_2019_CVPR} & 1/1 & 71.2 \\ \cline{2-4}
 & Group-Net**+BPAC~\cite{Zhuang_2019_CVPR} & 1/1 & 72.8 \\ \hline
\multirow{5}{*}{\tabincell{c}{{Faster-RCNN}\\{ResNet-50}\\{FCN-32s}}} & Full-precision~\cite{Zhuang_2019_CVPR} & 32/32 & 73.1 \\ \cline{2-4}
 & LQ-Nets~\cite{LQ-Net} & 3/3 & 70.7 \\ \cline{2-4}
 & Group-Net~\cite{Zhuang_2019_CVPR} & 1/1 & 67.2 \\ \cline{2-4}
 & Group-Net+BPAC~\cite{Zhuang_2019_CVPR} & 1/1 & 70.4 \\ \cline{2-4}
 & Group-Net**+BPAC~\cite{Zhuang_2019_CVPR} & 1/1 & 71.0 \\ \hline
\end{tabular}
\end{table}

\section{Future Trend and Conclusions}
\label{section5}
The binary neural networks based on 1-bit representation enjoy the compressed storage and fast inference speed, but meanwhile suffer from the performance degradation. To bridge the gap between the binary and full-precision models, as we summarized in this survey, there are various solutions proposed in recent years, which can be roughly categorized into the naive and the optimized. Our analysis shows that optimizing the binary network using different techniques can promise better performance. {These techniques, derived from different motivations, mainly focus on how to preserve the information in the forward propagation and how to optimize the network in the backward propagation. It shows that retaining the various information in forward and backward propagation is one of the key factors in training high-performance BNNs.}

Although much progress has been made, existing techniques for binary neural networks still face the performance loss, especially for the large network and datasets. The main reasons might include: (1) it is still unclear what kind of network structure is suitable for binarization, so that the information passing through the network can be preserved, even after binarization. (2) it is a difficult problem to optimize the binary network in a discrete space, even we have the gradient estimator or approximate function for binarization. 
We believe more practical and theoretical studies will emerge to answer the two questions in the future. 

Besides, as the mobile devices are becoming widely used in real world, more research efforts will be devoted to the applications to different tasks and deployment on different hardware. {For example, \cite{Wu2020Rotation} proposed a novel rotation consistent loss considering the open set characteristics of face recognition and achieves competitive performance using 4-bit compared to the full-precision model.} Therefore, there will arise the interesting topics such as customizing or transferring binary networks for different tasks, designing hardware-friendly or energy-economic binarization algorithms, \etc. 

{In addition to weights and activations, quantizing the backward propagation including gradients to accelerate the whole training process has arisen as a new topic recently. The unified framework proposed in \cite{zhu2019unified} proves the possibility of 8-bit training of neural networks from the accuracy and speed aspect. It is worthy to further explore the feasibility of binarized backward calculation for faster training time.}

{Last but not the least, the research on explainable machine learning indicates that there are critical paths in the prediction of neural networks and different network structures follow different patterns. So it is also meaningful to design mix-precision strategy according to the importance of layer and devise new architectures that are friendly to the information flow of binary neural networks.}

\section*{Acknowledgment}
This work was supported by National Natural Science Foundation of China (61872021, 61772057 and 61690202), Beijing Nova Program of Science and Technology (Z191100001119050), BSFC No. 4202039, and the support funding from State Key Lab. of Software Development Environment and Jiangxi Research Institute of Beihang University. 

\clearpage

\tiny
\bibliography{mybibfile}

\end{document}